\def\year{2021}\relax
\documentclass[letterpaper]{article} % DO NOT CHANGE THIS
\usepackage{aaai21}  % DO NOT CHANGE THIS
\usepackage{times}  % DO NOT CHANGE THIS
\usepackage{helvet} % DO NOT CHANGE THIS
\usepackage{courier}  % DO NOT CHANGE THIS
\usepackage[hyphens]{url}  % DO NOT CHANGE THIS
\usepackage{graphicx} % DO NOT CHANGE THIS
\usepackage{graphicx}  % DO NOT CHANGE THIS
\usepackage{natbib}  % DO NOT CHANGE THIS AND DO NOT ADD ANY OPTIONS TO IT
\usepackage{caption} % DO NOT CHANGE THIS AND DO NOT ADD ANY OPTIONS TO IT
\usepackage{subfig}
\nocopyright
\title{EasyRL: A Simple and Extensible Reinforcement Learning Framework}
\author {
  Neil Hulbert, Sam Spillers, Brandon Francis, James Haines-Temons, Ken Gil Romero, Benjamin De Jager, Sam Wong, Kevin Flora, Bowei Huang, Athirai A. Irissappane \\
}
\affiliations {
    University of Washington, Tacoma \\
    \{nhulbert, samds4, brandf3, jamesht, kgmr, benjad2, sam543, flora01, boweih, athirai\}@uw.edu
}

\begin{document}

\maketitle

\begin{abstract}
In recent years, Reinforcement Learning (RL), has become a popular field of study as well as a tool for enterprises working on cutting-edge artificial intelligence research. To this end, many researchers have built RL frameworks such as openAI Gym and KerasRL for ease of use. While these works have made great strides towards bringing down the barrier of entry for those new
to RL, we propose a much simpler framework called EasyRL,  
by providing an interactive graphical user interface for users to train and evaluate RL agents. As it is entirely graphical, EasyRL does not require programming knowledge for training and testing simple built-in RL agents. EasyRL also supports custom RL agents and environments, which can be highly beneficial for RL researchers in evaluating and comparing their RL models.
\end{abstract}

\section{Introduction}

Reinforcement Learning (RL) is a popular formalism for automated decision-making. It is a growing field with impressive advances. However, existing RL techniques have been mostly applied and evaluated on games~\cite{mnih2015human,silver2016mastering}. Apart from a small number of real-world applications~\cite{zeng2016online,moriyama2018reinforcement} that use RL solutions, RL is not a widely popular decision-making model across all domains. RL training, in general, is a cumbersome process. Several issues including (but not limited to) hyper-parameter tuning, sample efficiency, and training stability~\cite{schulman2017proximal} need to be carefully addressed during the training process. It has therefore become indispensable to have a solid background knowledge in RL and sufficient software development skills to successfully develop and train RL agents. The above requirements restrict the RL audience to only a handful of researchers who are experts in RL. Although RL can solve decision-making problems in other domains such as healthcare, transportation, networking, etc., people in these fields may find it cumbersome to train RL agents without sufficient expertise. 

Our main goal is to build a RL framework that can be used by diverse audiences from different domains. To this end, we propose the EasyRL framework for both native as well as non-native RL users to easily develop, train, and evaluate RL agents. The existing RL frameworks Keras-RL, Tensorforce, Horizon, HuskaRL, SimpleRL, AI-ToolBox, and Coach provide a range of built-in deepRL (deep reinforcement learning) techniques and some of them, such as Coach, create  visualization of the training process for debugging purposes~\cite{winderwebsite}. However, they do not support a user-friendly Graphical User Interface (GUI). Even the popular OpenAI gym only supports a range of environments to evaluate RL agents. Further, all the existing frameworks require some amount of RL and programming knowledge.

Our EasyRL framework offers an interactive GUI to build, train, and evaluate RL agents. It hosts a number of built-in RL agents (algorithms) and environments. Additionally, a user may develop his own custom RL agent or environment and add it to the framework. The EasyRL framework requires no (or minimal) programming skills for simple RL training. This differs from many previously existing RL frameworks by greatly simplifying access and reducing the technical barrier of entry to training RL agents. As the amount of personal computing resources has tremendously increased, the applicability of RL techniques to well-defined environments can be better leveraged to allow non-native RL users to themselves train RL agents. By introducing the user to RL and giving them the tools to create agents and environments themselves, our framework will improve the visibility as well as applicability of RL across different domains.

\begin{figure}[ht!]
\centering
    \includegraphics[keepaspectratio, height=3cm]{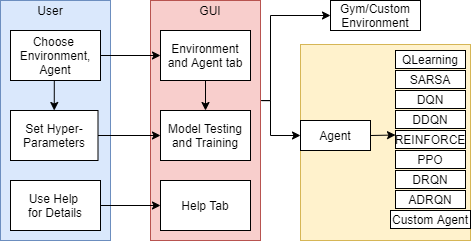}
    \caption{Structure of the EasyRL Framework}
    \label{fig:structure}
\end{figure}

\begin{figure*}[t]
    \centering
    \subfloat[Choosing Environment \& RL Agent ]{ \includegraphics[width=0.48\linewidth, height=6cm]{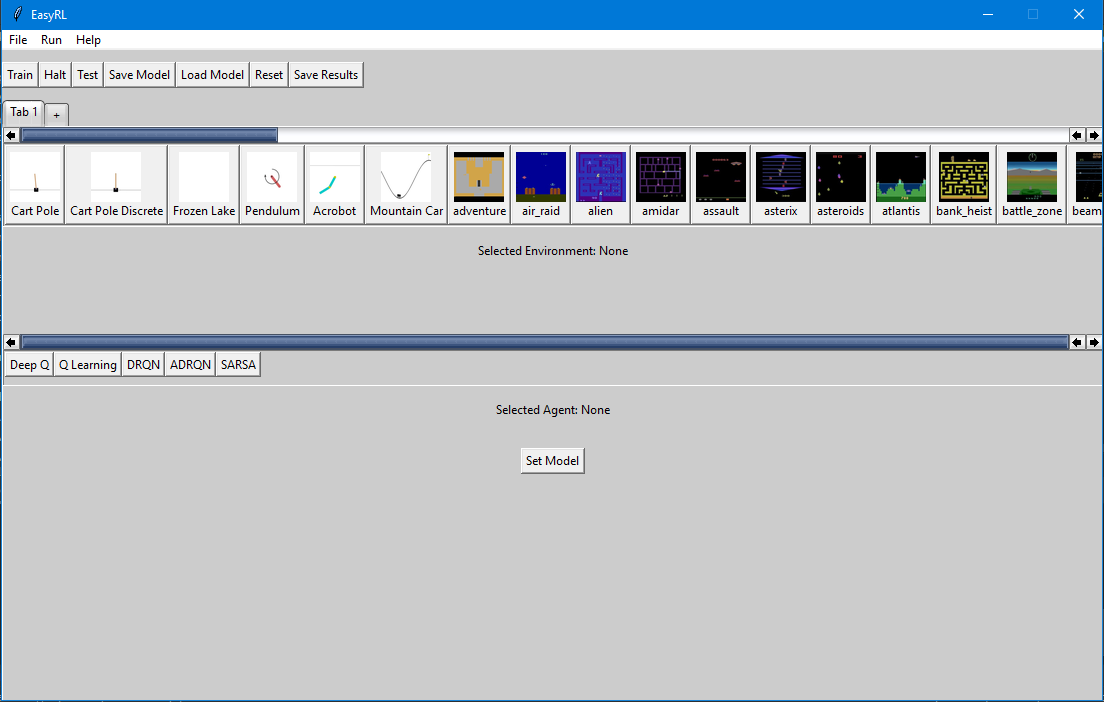} \label{fig:gui1}}%
    \hspace{2mm}
    \subfloat[Setting Hyper-Parameters \& Visualization]{\includegraphics[width=0.48\linewidth, height=6cm]{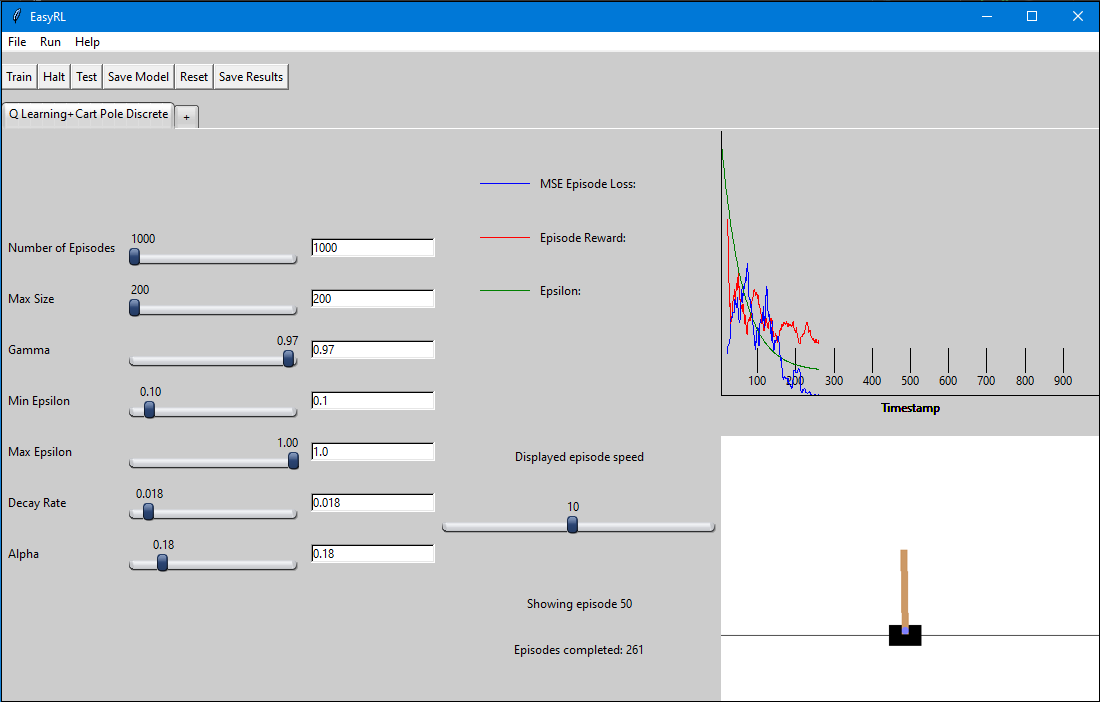} \label{fig:gui2}}%
    \caption{EasyRL Framework GUI}%
    \label{fig:gui}%
      \vspace{-2mm}
\end{figure*}

\section{The EasyRL Framework}
% \textbf{Explain GUI-1} \\

The proposed EasyRL framework allows the training and evaluation of RL agents on a variety of openAI gym as well as custom real-world environments. EasyRL follows a highly modularized implementation with abstractions such as \textit{Agent} and \textit{Environment}. The sequence diagram for navigating through the framework is shown in Fig.~\ref{fig:structure}. The GUI for the framework is shown in Fig.~\ref{fig:gui}. The user can select from a variety of RL agents and environments (see Fig.~\ref{fig:gui1}). The user can then set the hyper-parameters for training (see Fig.~\ref{fig:gui2}). The training results are plotted using metrics such as mean rewards and training loss. The graphs also show the epsilon annealing process. The training environment is dynamically rendered on the screen. The rendering speed can also be changed (see \textit{Display Episode Speed} in Fig.~\ref{fig:gui2}).

The trained RL model as well as results can be saved for future use. The user can load a previously trained RL agent using \textit{Load Model} and run test cases as well as visualize the results. The framework provides options to create custom RL agents and environments using \textit{Load Agent} and \textit{Load Environment}. We provide a detailed help guide to assist the user with these commands along with tooltip texts. %We also provide tool-tip texts for brevity. 

\begin{figure}[ht!]
\centering
    \includegraphics[keepaspectratio, height=3.5cm]{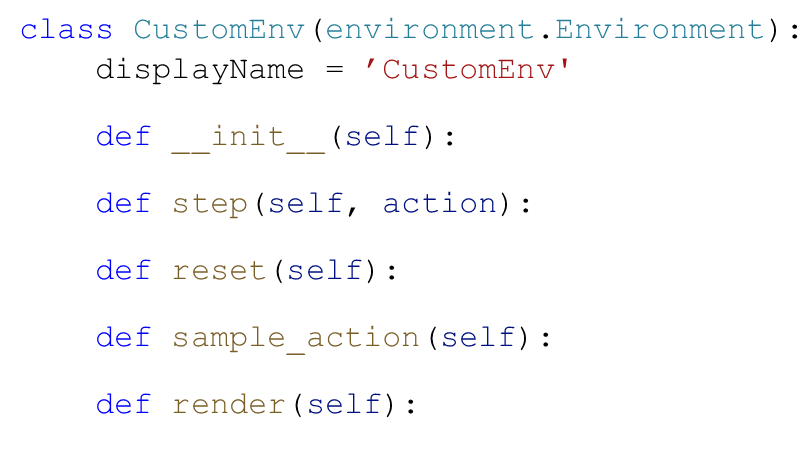}
    \vspace{-3mm}
    \caption{API for Custom Environment}
    \label{fig:cust}
       \vspace{-3mm}
\end{figure}

\subsection{RL Algorithms \& Environments}
EasyRL currently hosts a list of model-free algorithms that can handle both fully-observable environments such as Q-learning~\cite{watkins1992q}, SARSA~\cite{rummery1994line}, DQN~\cite{mnih2015human}, DDQN~\cite{van2016deep}, PPO~\cite{schulman2017proximal}, REINFORCE~\cite{williams1992simple} and partially-observable environments such as DRQN~\cite{hausknecht2015deep} and ADRQN~\cite{zhu2017improving}. The off-policy deepRL techniques mentioned above are implemented using standard experience replay for sampling experiences. It should be noted that our framework also supports model-based RL agents. The user is also allowed to create custom RL agents and import them to the EasyRL framework (as a python file).

The framework hosts a variety of OpenAI Gym environments (classic control and atari). The user can also create a custom environment by following the API shown in Fig.~\ref{fig:cust}. We have implemented some real-world environments, e.g., selecting sellers in e-markets~\cite{irissappane2014pomdp}. % and chemotherapeutic drug-dosage for cancer treatment~\cite{padmanabhan2017reinforcement}. 

The EasyRL framework is highly modularized and extensible (MVC design pattern). The EasyRL framework is predominately written in python and supports both tensorflow as well as pytorch deep learning libraries. EasyRL also supports C++ native implementations (see DRQNNative, DDQNNative) via CFFI which speeds up the training atleast by $5$ times. The framework, by default, uses local CPU/GPU during training, however, it can be easily configured to use resources remotely. Further, EasyRL supports training multiple RL agents in parallel via the python Threading library. The EasyRL framework is easy to install and is supported by linux, windows as well as iOS. We also provide a command-line interface, offering the same functionality as the GUI.

\subsection{Demo}
Our demonstration will show how a GUI can greatly simplify the process of developing, training, and testing a RL agent. We will demonstrate our simple installation procedure and show how a user with with minimal knowledge of RL and even programming can successfully train a RL agent. In addition to training and testing different combinations of agents and environments, we will show how to save and load pre-trained RL agents along with the results from a training or test run. We will demonstrate how to create custom environments and RL agents and show the training results for one such custom environment including the visualization graphs. Furthermore, we will show how multiple agents can be trained simultaneously and the improvement in training speed when native C++ implementation is used.

%\bibliography{ref}

\end{document}